# A Hybrid VDV Model for Automatic Diagnosis of Pneumothorax using Class-Imbalanced Chest X-rays Dataset

Tahira Iqbal[1], Arslan Shaukat[1], Usman Akram[1], Zartasha Mustansar[1], Yung-Cheol Byun[2]

[1]National University of Sciences and Technology (NUST), H-12, Islamabad, Pakistan

[2]Jeju National University, Jeju, South Korea

tahira.iqbal36@ce.ceme.edu.pk, arslanshaukat@ceme.nust.edu.pk, usman.akram@ceme.nust.edu.pk, zmustansar@rcms.nust.edu.pk, ycb@jejunu.ac.kr

**Abstract:**

Pneumothorax, a life threatening disease, needs to be diagnosed immediately and efficiently. The prognosis in this case is not only time consuming but also prone to human errors. So an automatic way of accurate diagnosis using chest X-rays is the utmost requirement. To-date, most of the available medical images datasets have class-imbalance issue. The main theme of this study is to solve this problem along with proposing an automated way of detecting pneumothorax. We first compare the existing approaches to tackle the class-imbalance issue and find that data-level-ensemble (i.e. ensemble of subsets of dataset) outperforms other approaches. Thus, we propose a novel framework named as VDV model, which is a complex model-level-ensemble of data-level-ensembles and uses three convolutional neural networks (CNN) including VGG16, VGG-19 and DenseNet-121 as fixed feature extractors. In each data-level-ensemble features extracted from one of the pre-defined CNN are fed to support vector machine (SVM) classifier, and output from each data-level-ensemble is calculated using voting method. Once outputs from the three data-level-ensembles with three different CNN architectures are obtained, then, again, voting method is used to calculate the final prediction. Our proposed framework is tested on SIIM ACR Pneumothorax dataset and Random Sample of NIH Chest X-ray dataset (RS-NIH). For the first dataset, 85.17% Recall with 86.0% Area under the Receiver Operating Characteristic curve (AUC) is attained. For the second dataset, 90.9% Recall with 95.0% AUC is achieved with random split of data while 85.45% recall with 77.06% AUC is obtained with patient-wise split of data. For RS-NIH, the obtained results are higher as compared to previous results from literature However, for first dataset, direct comparison cannot be made, since this dataset has not been used earlier for Pneumothorax classification.

## 1. INTRODUCTION

Pneumothorax can be interpreted as life-threatening condition which occur due to the collapse of respiratory system. The disease occurs as the air present inside the lungs is leaked to the space between chest walls and lungs. Due to this air, pressure is exerted on the lungs and it becomes difficult for the person to breath as the lungs cannot expand properly, thus the respiratory system collapse. Symptoms include shortness of breath and sudden pain in chest. In some cases these symptoms can be deadly, so it is very important to get it diagnosed in time [1]. One way of diagnosis is Radiographs and other is Chest Tomography (CT) scans. Because of cheap cost and availability of Chest X-ray (CXR) machines almost everywhere, doctors prefer to recommend X-rays instead of CT scan [2]. However identifying chest diseases from radiographs can be a challenging task for the radiologist. Hence, a computer aided diagnostic system is needed for the detection of Pneumothorax from the CXR images which can assist the radiologists.

Extensive research has been done in medical field after the emergence of Machine learning techniques including skin cancer detection [3], detection of arrhythmia [4] and detection of diabetic retinopathy [5]. Recently, detection of diseases from chest radiographs has become a hot topic. In [6] deep learning based frameworks have been proposed in order to detect lung nodules. An outstanding work has been done in [7] by proposing a 121-layered-Dense network in order to detect pneumonia with radiologist level performance along with predicting multiple thoracic diseases.

However, class imbalance is a massive problem in most of the medical images dataset [8]. In literature, mostly resampling techniques have been used, which could be disadvantageous as it can lead to removal of samples which might be important for training the model or it may lead to overfitting. So there is a need to propose a model which tackles the class imbalance problem efficiently along with predicting the presence of disease. In this research we aim to find an automated way to detect pneumothorax from images of chest x-rays by utilizing deep learning techniques along with comparing different existing approaches to tackle the class imbalance issue. Our proposed approach has been tested on two publicly available chest X-ray datasets.

The remaining paper is arranged as follows. Existing work for the detection of pneumothorax from chest x-rays and the different approaches in literature for the class-imbalance issue experimented in our research are discussed in Section 2. Section 3 describes the proposed VDV ensemble model for the pneumothorax detection. Section 4 describes the datasets used for our research purpose. Explained in Section 5 are the results and discussion. In Section 6, conclusion and future work are mentioned.

## 2. LITERATURE REVIEW

As stated earlier, several Artificial Intelligence (AI) based techniques have been implemented for the segmentation and automatic diagnosis of lung diseases. In [9] the researchers used a local dataset containing 32 pneumothorax x-ray images and 52 normal cases and identified the presence of pneumothorax by extracting features with Local binary pattern and using SVM as classifier. The mean accuracy achieved was 82%. In [10] comparison of two different Machine Learning approaches, bag of features (BoF) and CNN, was made with the intention of detection of normal and abnormal lung patterns related to 5 different pathologies, including identification of normal and pneumothorax from X-ray images. The experiments were performed on animal X-ray images. Total of 78 images were used for the pneumothorax detection and it was found that features extracted from CNN architecture outperformed BoF in all 5 different cases. Park *et al.* [11] trained YOLO Darknet-19 pre-trained model for automatically diagnosing pneumothorax, using a dataset containing 1596 Pneumothorax and 11137 normal X-ray images, which were acquired from tertiary hospitals. In [12] classification and localization was performed using a dataset of 1003 DICOM images, and comparison of CNN, Fully Convolutional Networks (FCN) and Multiple Instance Learning (MIL) technique was made, with CNN outperforming other techniques. In [13], chest CT scans were used for automatic detection of pneumothorax, where total of 280 chest CTs were used for training a CNN and then SVM was used as classifier. Thoracic

Ultrasound images were used in [14] to train a model for distinguishing between normal and pneumothorax cases. Image preprocessing techniques were implemented for the removal of textural information from the Ultrasound images and for image enhancement. Model accuracy was increased by the application of transfer learning and fine tuning techniques on pre-trained CNN architectures. Pixel classification based CNN approach was used for pneumothorax detection using a training set of 117 CXRs, achieving AUC value of 95% on a test set of 86 CXRs [15]. Texture analysis based technique was combined with supervised learning technique (KNN) in order to detect pneumothorax and this proposed framework was tested on a dataset of 108 CXRs giving performance of 81% and 87% in terms of sensitivity and specificity respectively. [16]. Jakhar *et al* [17] used the Kaggle pneumothorax dataset, for the purpose of segmentation of region of interest in chest x-ray images, while making use of U-Net architecture with ResNet as backbone.

From the literature, it has been observed that in most cases, dataset is either small or there exist a problem of class imbalance i.e. unequal number of images of different diseases. As per our findings, in most of the research carried out for the automatic detection of pneumothorax using an imbalance dataset, single approach have been used to resolve the problem of class imbalance as in [18]. Similarly in [19], the researchers proposed Under-Bagging based ensemble model for the said problem, where several subsets of training sets were created which were combined with minority class samples. Salehinejad *et al* [20] in their research used GANs with the aim of increasing the minority class samples.

These resampling methods (i.e. under-sampling and over-sampling) can become a reason of loss of important data or overfitting. Comparison of multiple approaches to solve class imbalance had been made by some of the researchers using commonly available datasets like MNIST, CIFAR etc. Buda *et al* [21] compared the performance of CNNs using multiple approaches including oversampling, under-sampling and thresholding, and evaluated their results on MNIST and CIFAR-10 dataset. However to the best of our knowledge, such comparison has not yet done using a medical image dataset. In our research we have made a comparison of different existing approaches to tackle this problem using publicly available pneumothorax dataset, along with proposing an ensemble based framework for the automatic diagnosis of pneumothorax from the CXRs which has been tested on two openly available datasets.

**2.1. Comparison of different methods for imbalance dataset**

Mainly two different approaches are available for imbalance dataset. The first one, known as data level methods, deals with altering the original dataset such that each class contains same number of samples. The second one is classifier level method in which the algorithms are adjusted to solve the said issue [22]. Explained below are the different approaches experimented in our research.

*2.1.1. Weight Balancing (Classifier level method):* It is one of the classifier level technique to solve the said issue [23]. In this technique, whole training set as provided in the original dataset is used, however different weights are assigned to majority and minority class data. The class weights are assigned according to formula given below:

$$class\ weight = \frac{n_{samples}}{n_{classes} * np.bincount\ (y)} \tag{1}$$

In Equation (1), $n_{samples}$ refer to the total size of training set, total number of classes is represented by $n_{classes}$, and $np.bincount(y)$ is a function which counts the frequency of each element in *y* array (i.e. count the frequency of 0 and 1 class elements separately).

*2.1.2. Under sampling (Data level method):* As the name suggests, a subset of samples is randomly selected from majority class so that we have equal number of observations in both classes [24]. Although in this approach enormous number of samples are discarded, still it has been found that in some situations, under-sampling works better than other approaches [25].

*2.1.3. Oversampling (Data level method):* In this approach, the sample size of minority class is increased so that it becomes equal to the majority class' sample size [26]. Some of these oversampling techniques include SMOTE [27], Cluster based oversampling [28] and DataBoost-IM [29]. In case of image dataset, another way to generate more samples is Data Augmentation explained in [30], as adopted for our experimentation purpose. This way more images can be generated for minority class making it equal to majority class samples.

*2.1.4. Ensemble (Hybrid approach):* This approach combines multiple techniques from both or one of the above-mentioned approaches. In case of using under-sampling method EasyEnsemble and BalanceCascade are used to train multiple classifiers [31]**.** Data level ensemble approach in [32] describes the way to create subsets of whole data, assuming identical distribution of observations, thus creating subsets of data which contain the same ratio of samples in each class as present in original dataset. The ensemble model experimented in our case finds its root from **[**33**]** which utilizes the idea of the data-level-ensemble with little modification. According to this, subsets of majority class are created in such a manner that the sample size of each subset of majority class is the same as the total sample size of minority class. These class-balanced (i.e. equal sample size in every class) subsets of training data (containing a subset of majority class combined with all samples of minority class) can then be utilized for classification purpose. The ensemble approach experimented for the sake of comparison of existing approaches utilizes VGG-16 as feature extractor from each subset and Linear SVM as classifier, while final output is based on voting method [34].

## 3. MATERIAL AND METHODOLGY

This section explains the proposed framework along with describing the feature extractors and the classifier used in this research. Section 3.1 explains the different CNN architectures used in our experiments. Section 3.2 explains the SVM classifier. Section 3.3 explains the proposed VDV model for the classification of pneumothorax from CXR images.

## 3.1. Convolutional Neural Networks (CNN)

Single or multiple convolutional layers are arranged in a particular manner with the aim of creating a neural network named as Convolutional Neural Network (CNN). CNN requires huge amount of data to train itself which can then be used for supervised or unsupervised decision making. It does so by extraction of features from the input data and adjusting weights of the neurons by forward and back propagation [35]. There are many different CNN architectures available, trained on ImageNet dataset and their weights can be used as initial weights for any classification problem.

In any CNN architecture, the convolutional layers are used for feature extraction from the input while the last fully-connected (FC) dense layers act as classifier. One of the ways to utilize the pre-trained CNNs is known as 'Fixed Feature extractor'. In this method the CNNs trained on large datasets, like ImageNet, are used as feature extractor by removing the last fully connected layers and features are extracted from the remaining CNN architecture. The extracted features can be fed to any classifier like SVM or softmax classifier [36 , 37]. As we have not trained the CNNs on our dataset, so we have not specified the training options like learning rate, optimizer and number of epochs. However Section 4.2 describes the number of extracted features from each CNN architecture and the number of layers from which features are extracted. The pre-trained CNN architectures that we have selected for our experimentation purpose are explained below.

*3.1.1. VGG-16:* One of the CNN architecture proposed by Simonyan *et al*. The architecture is composed of sixteen layers which includes twelve convolutional layers. These layers are predecessor of three fully-connected dense layers. The convolutional layers use 3×3 filters, stride and padding of 1. Followed by some of the convolutional layers is the 2×2 maximum pooling layer (stride of 2). There are 4096 neurons each in the first two dense layers. The third layer is meant for classification thus it contains 1000 channels. After the fully connected dense layers, there is a soft-max activation layer. This CNN architecture takes RGB image as input with default size of 224×224. In VGG-16, the total number of parameters is 14,714,688 [38]. The breakdown structure of VGG-16 architecture is shown in Table 1.

*3.1.2. VGG-19:* This model is the extension of VGG-16 except that it is comprised of 19 layers, out of which there are 16 convolutional layers and three FC layers. The architecture arrangement is same as VGG-16. There are 20,024,384 parameters in VGG-19. Like VGG-16, it takes 224×224 RGB image as its input. In order to use this architecture for classification problem, the last fully connected dense layer with 1000 neurons/channels is replaced by a dense layer containing neurons equal to the number of classes in the classification problem [38, 39]. The VGG-19 architecture is shown in Table 2.

**Table1** VGG-16 architecture

| Layer | Operation |
|---|---|
| Input layer ||
| Convolution | $[3x3\ conv]\ x\ 2$ |
| Pooling | $2x2\ max\ pool, stride\ 2$ |
| Convolution | $[3x3\ conv]\ x\ 2$ |
| Pooling | $2x2\ max\ pool, stride\ 2$ |
| Convolution | $[3x3\ conv]\ x\ 3$ |
| Pooling | $2x2\ max\ pool, stride\ 2$ |
| Convolution | $[3x3\ conv]\ x\ 3$ |
| Pooling | $2x2\ max\ pool, stride\ 2$ |
| Convolution | $[3x3\ conv]\ x\ 3$ |
| Pooling | $2x2\ max\ pool, stride\ 2$ |

**Table2** VGG-19 architecture

| Layer | Operation |
|---|---|
| Input layer ||
| Convolution | $[3x3\ conv]\ x\ 2$ |
| Pooling | $2x2\ max\ pool, stride\ 2$ |
| Convolution | $[3x3\ conv]\ x\ 2$ |
| Pooling | $2x2\ max\ pool, stride\ 2$ |
| Convolution | $[3x3\ conv]\ x\ 4$ |
| Pooling | $2x2\ max\ pool, stride\ 2$ |
| Convolution | $[3x3\ conv]\ x\ 4$ |
| Pooling | $2x2\ max\ pool, stride\ 2$ |
| Convolution | $[3x3\ conv]\ x\ 4$ |
| Pooling | $2x2\ max\ pool, stride\ 2$ |

| Classification | $4096D\ fully\ connected$ $4096D\ fully\ connected$ $1000D\ fully\ connected$ $softmax$ | Classification | $4096D\ fully\ connected$ $4096D\ fully\ connected$ $1000D\ fully\ connected$ $softmax$ |
|---|---|---|---|

*3.1.3. DenseNet-121:* DenseNet-121 is a CNN architecture with 121 layers. It has total of four dense blocks and a transition layer is present between every consecutive dense block. Every dense block consists of many convolutional layers and the transition layers consist of batch normalization, a convolution layer with 1x1 kernel and an average pooling layer of size 2×2. At the end of the architecture, there is a fully connected layer with soft-max activation function. It has 1000 neurons referring to the total number of classes in the ImageNet dataset on which it is trained. It takes RGB image with default input size of 224×224. There are 7,037,504 parameters in DenseNet121. As opposed to traditional CNNs, here every layer has a connection with all the other layers and a direct access to loss functions and original input is given to every layer. Concatenation of the feature maps extracted from all the previous layers is done and fed as input to any layer, and all the subsequent layers get the feature maps of the current layer as input. This special design enhances the flow of information throughout the network and also minimizes the vanishing gradient problem [40]. The detailed structure of DenseNet-121 is shown in Table 3.

**Table 3** DenseNet-121 architecture

| **Layer** | **Operation** |
|---|---|
| Input layer ||
| Convolution | $[7x7\ conv]$ |
| Pooling | $3x3\ max\ pool, stride\ 2$ |
| Dense Block | $\begin{bmatrix} 1x1\ conv \\ 3x3\ conv \end{bmatrix} x\ 6$ |
| Transition Layer | $[1x1\ conv]$ $2x2\ average\ pool, stride\ 2$ |
| Dense Block | $\begin{bmatrix} 1x1\ conv \\ 3x3\ conv \end{bmatrix} x\ 12$ |
| Transition Layer | $[1x1\ conv]$ $2x2\ average\ pool, stride\ 2$ |
| Dense Block | $\begin{bmatrix} 1x1\ conv \\ 3x3\ conv \end{bmatrix} x\ 24$ |
| Transition Layer | $[1x1\ conv]$ $2x2\ average\ pool, stride\ 2$ |
| Dense Block | $\begin{bmatrix} 1x1\ conv \\ 3x3\ conv \end{bmatrix} x\ 24$ |
| Classification | $7x7\ average\ pool$ $1000D\ fully\ connected, softmax$ |

### 3.2. Support Vector Machine (SVM)

A machine learning algorithm that can be utilized for the purpose of classification as well as regression. Here mapping of data points takes place in *n*-dimensional feature space, where *n* refers to the total number of features. For classification, hyperplane is created in such a way that it best separates the two classes and maximizes the margin. In our work, we have used both the linear SVM and polynomial kernel SVM. In practice, SVM algorithm is implemented using a kernel. Data space containing input data points are transformed into higher dimensional space using kernel tricks. This is done to convert the non-separable classification problem into a separable problem. The linear kernel can be implemented as the normal dot product between $x$ and $x_i$, where $x$ is the input vector and $x_i$ refers to each support vector. It is implemented using the following equation:

$$f(x, x_i) = B(0) + sum\ (a_i * (x, x_i)) \qquad (2)$$

For every input, training data is used for calculating $B(0)$ and $a_i$, using the learning algorithm. The SVM with polynomial kernel can distinguish the non-linear input space. It is expressed as follows:

$$f(x, x_i) = 1 + sum\ (x * x_i)^d \qquad (3)$$

where *d* represents the degree of polynomial [41]

### 3.3. Proposed Model

Among the different approaches that we have experimented for solving the class imbalance problem, the data-level ensemble (i.e. ensemble of subsets of dataset) performs better than other approaches. The superiority of a data-level-ensemble, which is created by creating several class-balanced subsets of data is proven in [33]. The results in [42] also show that the model-level-ensemble created by training different classifiers separately and later combining the individual classifier's results either by averaging or voting method give better performance. Thus uniting these two ideas (i.e. model-level-ensemble and data-level-ensemble), we present a novel framework VDV which is model-level-ensemble of different data-level-ensembles. It utilizes three different CNN architectures as fixed feature extractor and polynomial kernel SVM as classifier [43]. The selected CNN architectures for feature extraction purpose are VGG16, VGG19 and DenseNet121, thus the proposed framework is named as VDV model. In all these CNN architectures last fully connected layers are removed and the features extracted by the architecture are sent to the classifier. Like any other machine learning based CAD systems, our proposed framework comprises of two parts, i.e. training and testing. The block diagram for the training module is shown in Fig 1. Basically the training process makes use of three different CNN architectures as shown in Fig 1a .Block A refers to VGG-16, block B refers to DenseNet121 and block C refers to VGG-19, working as feature extractors. Training set is sent to each block separately and each block generates three SVM trained models. These trained models are later used for predicting the class of test sample. The internal working of each block is same except for using different CNN architectures. Fig 1b explains the generic working of each block. As we are dealing with class-imbalance-data, so first we create mini-training set, for which we divide the majority class samples into multiple subsets. Here we have two classes, Normal (negative class or class 0), and Pneumothorax (positive class or class 1). In our dataset, class 0 has majority number of samples, so it is divided in such a way that each subset has *n* number of samples of class 0, where *n* is same as the minority class' sample size (i.e. class 1). This way we will have N subsets of class 0, where N is equal to the imbalance ratio between class 0 and class 1. As in the SIIM dataset, the imbalance ratio is around 3.49:1, thus we have 3 subsets of class 0 which are referred as subset1, subset2 and subset3 in Fig [1b]. Then each subset is combined with whole minority class (i.e. class 1) thus creating a mini-training set. Features are extracted from each mini-training set and sent to SVM classifier, which generates SVM trained model for every mini-training set (referred as set1, set2 and set3). Note that this process is repeated for all the three blocks which are shown in Fig [1a].

Fig [2] represents the block diagram of the test module. Here the trained SVM models generated by the three blocks in training module are used. Fig [2a] shows the workflow for predicting class of any sample. The test CXR image is sent to each block (Block A, Block B and Block C) which outputs the class prediction. As we have three blocks referring to different CNN architectures (i.e. VGG-16, DenseNet121 and VGG-19) as feature extractors, so each block generates its own prediction, these three predictions are used to make final decision based on Majority Voting method (i.e. maximum occurring predicted class is taken as final result). The internal generic working of each block of test module is shown in Fig [2b]. CNN architecture (with respect to each block) is used for extracting features from the test CXR image which are then sent to each of the three trained SVM models. Each SVM trained model predicts a class of test samples which are then combined together based on voting method.

In all the experiments, we keep the default input size of the images, i.e. 224×224. Note that for VGG-16 and VGG-19 we use all the 25088 extracted features in training and testing process, however because of really large number of features in case of DenseNet121 (i.e. 50176), we have to minimize the number of features else we cannot apply Kernel SVM because of memory issue. Principal Component Analysis (PCA) is applied for feature reduction, the total number of features is reduced to 4758. This number is calculated using Singular value decomposition (svd) solver method [43]. Moreover instead of 8296 samples, we keep 7137 samples of Normal class in training set, i.e. 3 times more than the number of samples in pneumothorax class. It is done in order to make class-balanced training subsets.

For authentication, we test our proposed framework on Random Sample of NIH Chest X-ray (RS-NIH) dataset as well. We select normal and pneumothorax samples from the dataset, thus we have an imbalance dataset with ratio of 11:1, i.e., for every sample of pneumothorax there are 11 samples of normal CXRs. It is important to mention here that in training set we keep 2376 samples of Normal class instead of 2435 samples, so that each mini-training set has equal number of normal and pneumothorax samples. The only difference while experimenting with this dataset is that, here we make 11 subsets of normal class samples instead of three subsets. Rest of the implementation is same as explained above.

The results on Test set for each separate Block (i.e. Block A, Block B, and Block C) along with our proposed VDV model on SIIM dataset and RS-NIH dataset are reported in Section [4].

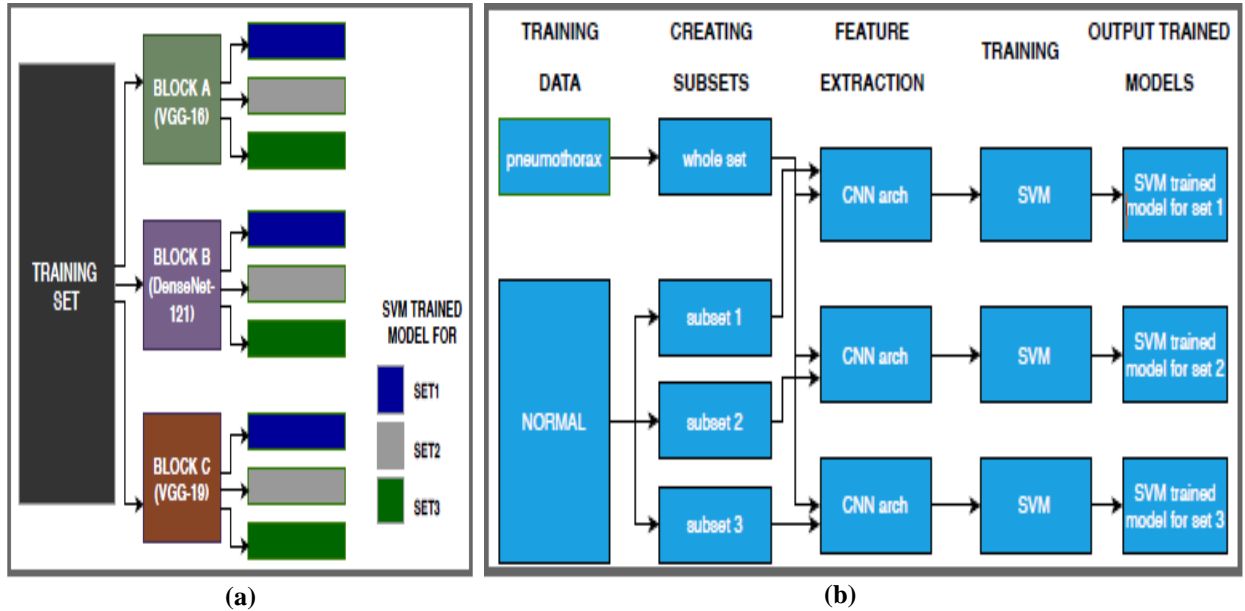

**Fig. 1** Training module of our proposed VDV model in shown above. a) Shows the Block diagram for the Training Module. Here each block uses different CNN architecture for feature extraction and outputs three trained SVM models with respect to each set. b) Shows the Internal Working of each Block in Training Module. After creating subsets of training data, features are extracted from each mini-training set using one of the three CNN architectures, then SVM model is trained on these extracted features separately, thus generating SVM trained model with respect to each mini-set of training data.

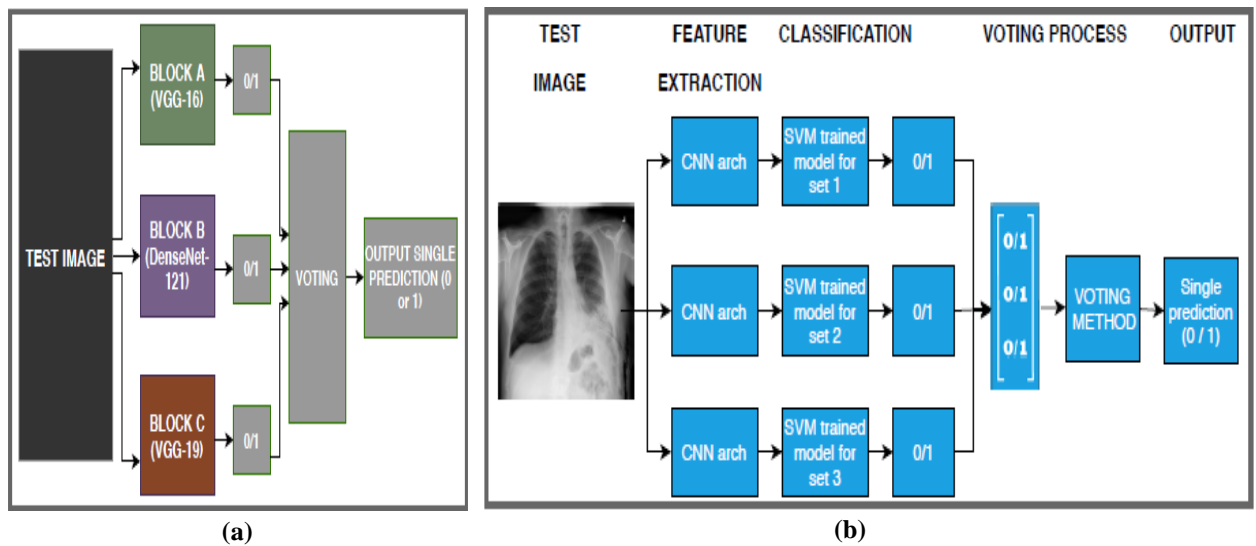

**Fig. 2** Test module of our proposed VDV model is shown above. a) Shows the Block diagram for Test module. Here each block uses different CNN architecture for feature extraction from test image and outputs predicted class. These predictions are sent to voting unit which outputs the maximum occurring class. b) Shows the Internal working of each Block in Test Module, in which trained SVM model with respect to each set, predicts the class of the test sample based on features extracted by CNN architecture. These predictions are combined using the Voting method to obtain a single prediction.

## 4. EXPERIMENTAL SETUP

### 4.1. Datasets

*4.1.1. SIIM-ACR Pneumothorax Dataset*: The first dataset selected for our experimentation purpose is available on Kaggle [44], which contains stage-1 training and testing data from "SIIM-ACR Pneumothorax Segmentation competition", in Portable Network Graphics (png) format. There are 12047 chest X-ray (CXR) images along with training and testing list. Training list contains 8296 normal CXR images and 2379 CXRs with pneumothorax, while the testing set contains 1082 normal CXRs and 290 images of the other class. The original size of X-ray images is 1024×1024. However, for our experimentation purpose we resize the images to 224×224. Main reason for selecting this dataset is that no classification results have been reported on this dataset before. Also it provides same number of RLE (Run Length Encoded) masks which can later be used for segmentation purpose. Table 4 summarizes the details of the dataset, where N represents Normal CXRs and P represent CXRs with pneumothorax. Some of the images from this dataset are represented in Fig 3.

**Table 4** Summarized details of SIIM dataset

| Resolution | | 1024 x 1024 |
|---|---|---|
| Dataset size | | 12047 |
| No of classes | | 2 |
| Training set | N | 8296 |
| | P | 2379 |
| Testing set | N | 1082 |
| | P | 290 |

*4.1.2. Random Sample of NIH Chest X-ray dataset (RS-NIH)*: The second dataset on which we have performed experiments using our proposed method is "Random Sample of NIH Chest X-ray Dataset" which is provided by National Institutes of Health NIH and is available on Kaggle [45]. The full NIH Chest X-ray-14 dataset (NIH-CXR) contains 112,120 images, with 15 classes, covering 14 different thoracic pathologies and 15[th] being the normal case. The "Random Sample of NIH Chest X-ray dataset (RS-NIH)" chosen for our research purpose is a sample version of the NIH-CXR dataset and contains 5% of the total number of samples, and each pathology is present in the same ratio as is present in full dataset. Each image has a resolution of 1024×1024. The sample dataset contains 3044 images of No-finding, Infiltration:967, Effusion:664, Atelectasis:508, Nodule:313, Mass:284, Pneumothorax:271, Consolidation:226, Pleural Thickening:176, Cardiomegaly:141, Emphysema:127, Edema:118, Fibrosis:84, Pneumonia:62 and 13 images of Hernia. For our experiments, we keep the images of No-finding case and pneumothorax samples. Just like the NIH-CXR dataset is divided into 80% training and 20% testing set, we also split our data into training and testing set with the same ratio. Note that two different protocols have been followed while splitting the dataset. First being a random split of data and second being a patient wise split, i.e. CXRs from the same patient can only be present in either training or testing set. The details of this dataset is given in Table 5.

**Table 5** Summarized details of RS-NIH

| Resolution | | 1024 x 1024 |
|---|---|---|
| Dataset size (with 14 classes) | | 5606 |
| No of classes chosen | | 2 |
| Training set | N | 2376 |
| | P | 216 |
| Testing set | N | 609 |
| | P | 55 |

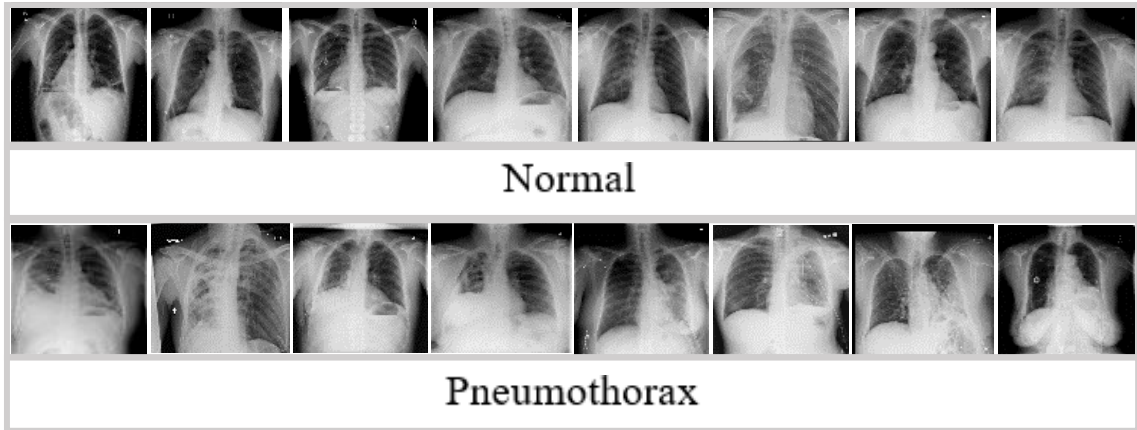

**Fig. 3** CXR images from the SIIM dataset

### 4.2. Experimental Settings

Keras with Tensorflow backend is used in our research with Python as programming language. Our research comprises of two parts, first one is to compare different existing approaches to tackle the imbalance problem and second one is to propose a framework (VDV) for automatic diagnosis of the disease, which is tested on two different datasets. In both the experiments, main task is features extraction and classification of CXR images as normal or pneumothorax. For the first part, i.e. comparison of techniques, we select pre-trained VGG-16 model (with ImageNet weight) as feature extractor based on its structural simplicity [35], with Linear SVM as classifier. For the proposed VDV network, three different CNN architectures are selected, VGG-16, VGG-19 and DenseNet121. The details of input and output sizes, number of parameters and number of layers in each architecture are summarized in Table 6. These pre-trained models with ImageNet weights are utilized for the purpose of extraction of features from the images. Last fully connected (FC) layers of these pre-trained models are removed as those are meant for classification purpose, instead we have used polynomial kernel SVM as classifier with gamma value 0.002 and C equal to 100. The values for kernel SVM are selected using grid search method.

Note that for the proposed framework we chose poly kernel SVM as it is a proven fact that SVM with poly kernel perform better than Linear SVM [46]. Moreover instead of using the last fully connected dense layers for classification, we chose SVM as it had been found to be more effective [47, 48].

Table 6 Parameters configuration of CNN

| CNN | Shape | | Features | Parameters | | Layers |
|---|---|---|---|---|---|---|
| | Input | Output | | Trainable | Non-Trainable | |
| **VGG-16** | 224×224×3 | 7×7×512 | 25,088 | 14,714,688 | 0 | 19 |
| **VGG-19** | 224×224×3 | 7×7×512 | 25,088 | 20,024,384 | 0 | 22 |
| **DenseNet-121** | 224×224×3 | 7×7×1024 | 50,176 | 6,953,856 | 83,648 | 427 |

### 4.3. Performance Measure

For evaluation of a model, selection of performance metrics is important. As our training as well as testing data is imbalanced, only accuracy is not a good performance measure [34] that is why we select Area under Receiver Operating Characteristic curve (AUC) and Recall as our performance metric. In addition to these, we also report the results with other performance metrics which include Accuracy, Specificity, Precision, Geometric mean (G-mean) [49], F1 and F2 score [50]. AUC is calculated by the area under Receiver Operating curve which is defined in terms of true positive rate and false positive rate [51]. In all the following expressions, TN, TP, FN and FP denotes True Negative, True Positive, False Negative and False Positive respectively. The expressions for calculating Accuracy, Recall, Precision and Specificity are given below:

$$Accuracy = \frac{TN + TP}{TN + FP + FN + TP} \quad (4)$$

$$Recall = \frac{TP}{TP + FN} \quad (5)$$

$$Precision = \frac{TP}{TP + FP} \quad (6)$$

$$Specificity = \frac{TN}{TN + FP} \quad (7)$$

The combination of recall and precision is an important metric known as F-score. It is calculated using $F_\beta$, where $\beta$ is assigned a different value, based on the problem statement. If the aim is to avoid misclassification of negative samples as positive ones, i.e. giving more importance to precision, then $\beta$ is assigned value equal to 0.5. However if it is intended to never miss positive class samples, like in our case, aim is to make a classifier which should avoid missing pneumothorax samples, i.e. giving more importance to Recall, then value $\beta$ is set to 2. If both precision and recall are given equal importance, then $\beta$ is assigned a value equal to 1. In our experiments, we have calculated $F_1$ and $F_2$ score by substituting $\beta$ as 1 and 2 respectively. The expression for $F_\beta$ and G-mean are given below:

$$F_\beta = (1 + \beta^2) \frac{Recall \times Precision}{(\beta^2 . Precision) + Recall} \quad (8)$$

$$G\ mean = \sqrt{Recall \times specificity} \quad (9)$$

# 5. EXPERIMENTAL RESULTS AND DISCUSSION

## 5.1. Results

The first part of our work where comparison of existing class imbalance approaches is performed, utilizes openly available SIIM Pneumothorax dataset. Accuracy, Recall/Sensitivity, Specificity and AUC for all the experiments have been reported in this section. Table 7 summarizes the results for different existing class imbalance approaches experimented in this research. Here Column 2 (i.e. No. of Training Samples) refers to the total number of CXR images in each class, used in every approach, separately. Based on highest AUC value achieved in case of ensemble model, which is 80.02%, it can be inferred that the data-level ensemble model outperforms other existing approaches for class imbalance issue. Moreover, it is evident that sensitivity value is highest of all in case of ensemble model which shows that maximum correct identification of the pathology is achieved using an ensemble model, compared to any other existing approach.

Table 7 Comparison of different existing approaches for class imbalance problem

| Technique | No of Training Samples | | ACC (%) | REC (%) | SPE (%) | AUC (%) |
|---|---|---|---|---|---|---|
| | **Normal** | **Pneumothorax** | | | | |
| **Weight balancing** | 8296 | 2379 | 79.08 | 48.96 | 87.15 | 78.8 |
| **Under-sampling** | 2379 | 2379 | 72.15 | 68.62 | 73.10 | 77.67 |
| **Over-sampling** | 8296 | 8296 | 77.7 | 50 | 85.20 | 77.76 |
| **Ensemble** | 2379 (in each subset) | 2379 (in each subset) | 75.22 | **79.65** | 74.09 | **80.02** |

*ACC: Accuracy, REC: Recall, SPE: Specificity*

Based on these results, we propose our framework named as VDV model, the detailed performance of which is summarized in Table 8. As our model is an ensemble of three data-level-ensembles using three different CNN architectures, so we have first reported the individual model results in first three rows of Table 8, while the last two rows show the performance of the proposed VDV model on SIIM and RS-NIH dataset respectively. Note that the individual model performance is reported for SIIM dataset only. The AUC value achieved by our framework on SIIM dataset is 86.0% and sensitivity value of 85.17% which shows that a model-level-ensemble of different data-level-ensembles gives far better results as compared to a single CNN architecture used in any data-level-ensemble.

Moreover, in case of random split of dataset, our proposed framework performs better for RS-NIH dataset while giving 95.0% AUC, 82.68% accuracy and 90.9% recall value. On the other hand, following a patient-wise data split, the proposed VDV model achieved AUC of 77.06%, accuracy of 69.12%, and recall value of 85.45%.

Table 8 Performance of proposed framework VDV

| | ACC (%) | REC (%) | SPE (%) | PREC (%) | F1 (%) | F2 (%) | G-mean (%) | AUC (%) |
|---|---|---|---|---|---|---|---|---|
| **SIIM DATASET** | | | | | | | | |
| **VGG-16** | 77.55 | 83.79 | 75.87 | 48.21 | 61.2 | 73.01 | 79.73 | 86±0.01 |
| **VGG-19** | 77.04 | 82.06 | 75.69 | 47.5 | 60.17 | 71.64 | 78.8 | 86± 0.01 |
| **DenseNet-121** | 76.32 | 80.68 | 75.04 | 46.4 | 58.94 | 70.31 | 77.81 | 85±0.00 |
| **VDV (SIIM)** | 78.27 | 85.17 | 76.43 | 49.2 | 62.37 | 74.3 | 80.68 | 86±0.00 |

| RS-NIH DATASET (RANDOM SPLIT) | | | | | | | | |
|---|---|---|---|---|---|---|---|---|
| VDV(RS-NIH) | 82.68 | **90.9** | 81.93 | 31.25 | 46.5 | 65.78 | 86.3 | **95 ±0.01** |
| RS-NIH (WITH PATIENT WISE SPLIT) | | | | | | | | |
| VDV (RS-NIH) | 69.12 | **85.45** | 67.65 | 19.26 | 31.43 | 50.64 | 76.03 | **77 ±0.06** |

*PREC: Precision, F1: $F_1$ score, F2: $F_2$ score.*

The model performance in terms of AUC is shown in Fig 4a and 4b. Fig 4a represents the performance of our proposed model on SIIM dataset, where the ROC curves for individual data-level-ensemble utilizing single CNN architecture along with ROC curve of VDV model on SIIM dataset are plotted. The performance of our proposed framework in terms of AUC on RS-NIH dataset is represented in Fig 4b.

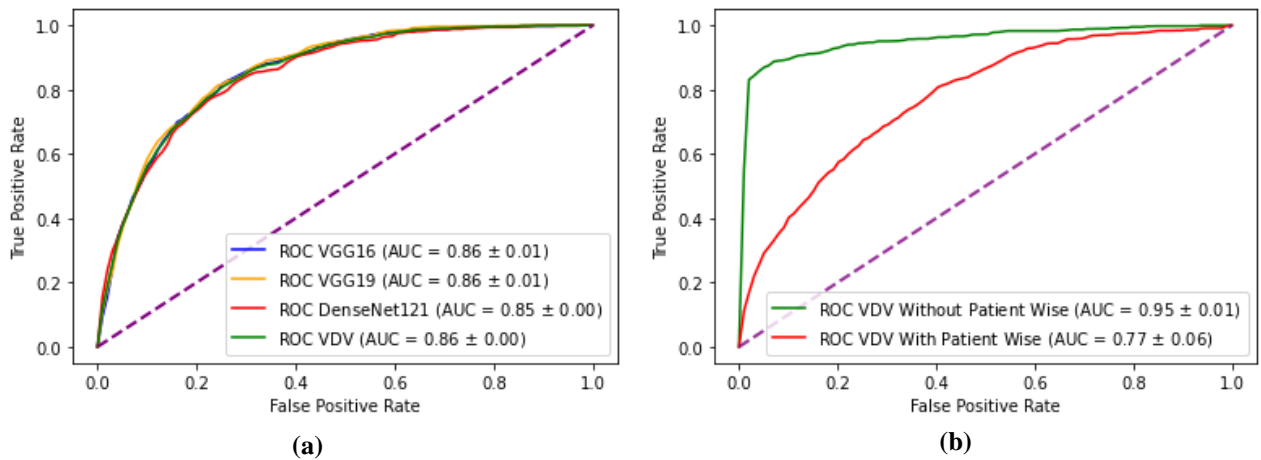

**Fig. 4** AUC plot for proposed VDV model. a) On SIIM dataset. b) On RS-NIH dataset.

The confusion matrix for the performance of VDV model on the SIIM and RS-NIH test set (with random split and patient wise split) are shown in Table 9 and Table 10 respectively. For the SIIM pneumothorax dataset our proposed model correctly identifies 247 pneumothorax cases while 43 are misclassified, and the total number of correctly classified Normal CXRs is 827 while 255 are misclassified as pneumothorax. For the RS-NIH dataset, with random split of data, 50 out of 55 samples are correctly classified as pneumothorax while 499 out of 609 samples are correctly identified as Normal CXRs. For patient-wise data split of RS-NIH dataset, 47 out of 55 and 412 out of 609 samples are correctly classified as pneumothorax and Normal x-rays respectively.

**Table 9** Confusion Matrix for SIIM dataset

| | **Predicted Class** | |
|---|---|---|
| **Actual Class** | Normal | Pneumothorax |
| Normal | 827 | 255 |
| Pneumothorax | 43 | 247 |

**Table 10** Confusion Matrix for RS-NIH

| Random Split of dataset | | |
|---|---|---|
| | **Predicted Class** | |
| **Actual Class** | Normal | Pneumothorax |
| Normal | 499 | 110 |
| Pneumothorax | 05 | 55 |

| Patient-wise Split of dataset | | |
|---|---|---|
| | **Predicted Class** | |
| **Actual Class** | Normal | Pneumothorax |
| Normal | 412 | 197 |
| Pneumothorax | 08 | 47 |

The comparison of different existing researches on detection of pneumothorax with our proposed model for SIIM dataset is provided in Table 11. The total number of normal and pneumothorax CXR used by each researcher in training and testing set is also given. The sub-column referred as B in the table shows if the dataset is class-balanced or imbalance in nature. The last column depicts if the dataset used is publicly available or not. Although greater value in terms of Sensitivity (Sen) and Accuracy (ACC) is achieved by other authors, however it can be clearly seen that the imbalance ratio of their test set is relatively small and also, the test set contains very few images compared to our test set. Moreover, they have used private dataset while dataset that we have used is publicly available, hence other researchers can add to this work.

**Table 11** Comparison of pneumothorax classification models for SIIM Dataset

| Author | Training Set | | | Testing Set | | | Results | | | | Dataset public |
|---|---|---|---|---|---|---|---|---|---|---|---|
| | N | P | B | N | P | B | ACC (%) | REC (%) | SPE (%) | AUC (%) | |
| Chan [9] | 36 | 22 | ✘ | 16 | 10 | ✘ | 82.20 | --- | --- | --- | ✘ |
| Yoon [10] | 24 | 24 | ✓ | 15 | 15 | ✓ | 96.60 | 100.0 | 93.8 | --- | ✘ |
| Park [11] | 10887 | 1343 | ✘ | 250 | 253 | ✓ | --- | 89.7 | 96.4 | 98.4 | ✘ |
| Gooben [12] | 350 | 453 | ✘ | 87 | 113 | ✘ | --- | --- | --- | 96.2 | ✘ |
| Li Xiang [13] | 30 | 50 | ✘ | 40 | 160 | ✘ | 96.50 | 100.0 | 82.5 | --- | ✘ |
| Proposed model (VDV) | 8296 | 2379 | ✘ | 1082 | 290 | ✘ | 78.27 | 85.17 | 76.43 | 86.0 | ✓ |

*N: No. of Normal CXRs, P: No. of CXRs with Pneumothorax, B: Balance or Imbalance Dataset.*

For the RS-NIH dataset, we can directly compare our results with [52] in which same RS-NIH dataset has been used for classification purpose, as presented in Table 12. Note that in [52], multi-label classification was performed considering 14 different chest diseases, so we have reported the achieved AUC value for pneumothorax classification.

**Table 12** Comparison of pneumothorax classification models for RS-NIH Dataset

| Author | Data-set | Description | Results (%) |
|---|---|---|---|
| Modal[52] | RS-NIH | Multi-label classification of 14 thoracic diseases | AUC= 54.0 |
| **Random split of data** | | | |
| **Proposed model** | RS-NIH | Binary classification (normal and pneumothorax CXRs ) | **ACC=82.68**<br>**AUC=95.0** |
| **Patient-wise split of data** | | | |
| **Proposed model** | RS-NIH | Binary classification (normal and pneumothorax CXRs ) | **ACC=69.12**<br>**AUC=77.06** |

*RS-NIH: Random Samples of NIH Chest X-ray dataset*

## 5.2. Discussion

Our work in this paper comprises of two parts, so we will discuss the result of each part separately. The fact that data-level-ensemble outperforms other existing approaches for class imbalance problem is because it makes use of whole dataset in such a way that training is done piece-wise using class-balanced subsets of whole data. It is generally observed that a balanced dataset performs better than class-imbalance data, so training the classifier on class-balanced subsets of dataset and then combining their results gives better performance as compared to other techniques. Now as a single model based data-level-ensemble (i.e. data-level ensemble with a single CNN architecture as feature extractor) gives quite good results, we have designed a model-level-ensemble of three data-level-ensembles, with each data-level ensemble using different CNN architecture as feature extractor. Utilizing three different CNN architectures in three data-level-ensembles separately allows our model to give better performance and utilization as opposed to the previously used single data-level-ensemble models.

In the work done so far, in field of class imbalance problem, where comparisons among different approaches are presented, just like our results proves, single architecture based data-level-ensembles have outperformed compared to other existing approaches. However, most of the previous works in literature have reported the results using MNIST or CIFAR dataset. Moreover, instead of proposing a new approach, most of the researchers using imbalanced medical images dataset have used a single approach, either oversampling or under sampling. We have not only made a comparison of different existing approaches using our real-life medical images dataset, but have also presented a novel framework which finds its roots from the concepts of data-level and model-level ensemble. To our knowledge, this type of ensemble model has never been proposed which not only solves the issue of class imbalance while using whole imbalanced dataset, but also takes advantage of the performance of different CNN architectures.

The comparison of our proposed model with existing literature is provided in Table 11 and Table 12. In Table 11, the comparison of performance of our proposed VDV model on SIIM datasets with existing work is presented, where it can be seen that although better results in terms of accuracy and AUC are achieved by other authors, but they have used comparatively smaller test dataset, and mostly datasets are balanced or have minimal ratio of imbalance, while our dataset is imbalance in terms of both training and testing set. Moreover, their datasets are not publicly available.

In addition, we have also tested our proposed framework on openly available RS-NIH dataset and the comparison with existing work is provided in Table 12. We have referred and compared our results with the paper in which pneumothorax was considered as a separate class in a multi-label classification problem, while we have not considered the papers in which the same dataset was used in order to differentiate between normal and abnormal CXRs without considering any specific pathology. In Table 12, it can be clearly seen that the results obtained by the proposed VDV model using both the data-split protocols (i.e. random split and patient-wise data split) are far better as compared to those achieved in [52], where random split of data was considered. Note that the higher performance of VDV model in case of random split of data as compared to patient-wise split is because in case of random split of dataset, there are chances that Chest X-rays from same patient might be present in both training and testing set, whereas in patient-wise split there is no such overlap.

The limitation of this proposed framework is that it cannot be experimented with K-fold cross validation, because it requires large number of subsets to be created based on the imbalance ratio. So in case of bigger dataset with large number of samples or highly class imbalance datasets, it would be computationally expensive and complex to perform K-fold cross validation using VDV model.

In the end, we can say that our work surpasses previous works performed till date in field of pneumothorax, as we have used publicly available datasets with the aim to allow other researchers to study, comprehend and offer their input. Our results prove that the proposed VDV framework (i.e. model-level-ensemble of data-level-ensembles) performs better than the existing approaches and can be used for any class-imbalance dataset.

## 6. CONCLUSION

Pneumothorax can be a deadly disease and there is a need to correctly identify it in time. With the advancement in deep learning technology, and its ability to make unsupervised, wise decisions, an efficient automatic diagnostic system can be proposed for detection of pneumothorax. For proposing such a framework for automatic detection of the diseases using a highly imbalance data, we have first analyzed different techniques for class imbalance problem using a real life medical image dataset. After finding out that data-level-ensemble (i.e. ensemble of subsets of training data) performs best of all, we have presented a model by combining the ideas of ensemble of models and ensemble of data. Our results have shown that this doubly ensemble VDV model outperforms single data-level-ensemble model with a single CNN architecture as feature extractor. Results reported on SIIM dataset in our paper will serve as baseline, since we are the first one to use this dataset for classification. Our achieved results on RS-NIH dataset are higher which also validates the performance of our proposed framework. So one can use our proposed framework for any imbalanced dataset with a little modification in terms of feature extractor CNN architecture and image input size. In future, we can propose utilization of this framework for bigger datasets, for example full NIH Chest X-ray-14 dataset. Also, a segmentation model using SIIM dataset can be developed which will be more helpful for the radiologists in correctly identifying the disease.

# Declarations

### Conflict of Interest
The authors have no conflict of interest to disclose.

### Funding
No funding was received to carry out the research.

### Availability of Data and material
Both the datasets used in this study are available on Kaggle and the links are mentioned in the manuscript.

# REFERENCES

[1] *Pneumothorax*. Harvard Health Publishing. (2019). Retrieved 12 June 2020, from http://www.health.harvard.edu/a_to_z/pneumothorax-a-to-z.

[2] Qin, C., Yao, D., Shi, Y., & Song, Z. (2018). Computer-aided detection in chest radiography based on artificial intelligence: a survey. *Biomedical engineering online*, *17*(1), 113. doi: 10.1186/s12938-018-0544-y.

[3] Esteva, A., Kuprel, B., Novoa, R. A., Ko, J., Swetter, S. M., Blau, H. M., & Thrun, S. (2017). Dermatologist-level classification of skin cancer with deep neural networks. *nature*, *542*(7639), 115-118. doi: 10.1038/nature21056.

[4] Rajpurkar, P., Hannun, A. Y., Haghpanahi, M., Bourn, C., & Ng, A. Y. (2017). Cardiologist-level arrhythmia detection with convolutional neural networks. *arXiv preprint* arXiv: 1707.01836.

[5] Gulshan, V., Peng, L., Coram, M., Stumpe, M. C., Wu, D., Narayanaswamy, A., ... & Kim, R. (2016). Development and validation of a deep learning algorithm for detection of diabetic retinopathy in retinal fundus photographs. *Jama*, *316*(22), 2402-2410. doi:10.1001/jama.2016.17216.

[6] Huang, X., Shan, J., & Vaidya, V. (2017, April). Lung nodule detection in CT using 3D convolutional neural networks. In *2017 IEEE 14th International Symposium on Biomedical Imaging (ISBI 2017)* (pp. 379-383). IEEE. doi: 10.1109/ISBI.2017.7950542.

[7] Rajpurkar, P., Irvin, J., Zhu, K., Yang, B., Mehta, H., Duan, T., & Lungren, M. P. (2017). Chexnet: Radiologist-level pneumonia detection on chest x-rays with deep learning. *arXiv preprint* arXiv:1711.05225.

[8] Vasudevan, H., Michalas, A., Shekokar, N., & Narvekar, M. (2020). *Advanced computing technologies and applications* (p. 300). Singapore: Springer. doi: 10.1007/978-981-15-3242-9.

[9] Chan, Y., Zeng, Y., Wu, H., Wu, M., & Sun, H. (2018). Effective Pneumothorax Detection for Chest X-Ray Images Using Local Binary Pattern and Support Vector Machine. *Journal Of Healthcare Engineering*, *2018*, 1-11. doi: 10.1155/2018/2908517.

[10] Yoon, Y., Hwang, T., & Lee, H. (2018). Prediction of radiographic abnormalities by the use of bag-of-features and convolutional neural networks. *The Veterinary Journal*, *237*, 43-48. doi: 10.1016/j.tvjl.2018.05.009.

[11] Park, S., Lee, S. M., Choe, J., Cho, Y., & Seo, J. B. (2019, January). Performance of a deep-learning system for detecting pneumothorax on chest radiograph after percutaneous transthoracic needle biopsy. *European Congress of Radiology, 2019*. doi:10.26044/ecr2019/C-0334.

[12] Gooßen, A., Deshpande, H., Harder, T., Schwab, E., Baltruschat, I., Mabotuwana, T., Cross, N., & Saalbach, A. (2019). Deep Learning for Pneumothorax Detection and Localization in Chest Radiographs. *arXiv preprint* arXiv:1907.07324.
[13] Li, X., Thrall, J. H., Digumarthy, S. R., Kalra, M. K., Pandharipande, P. V., Zhang, B., ... & Li, Q. (2019). Deep learning-enabled system for rapid pneumothorax screening on chest CT. *European journal of radiology*, *120*. doi: 10.1016/j.ejrad.2019.108692.
[14] Lindsey, T., Lee, R., Grisell, R., Vega, S., & Veazey, S. (2018, November). Automated pneumothorax diagnosis using deep neural networks. In *Iberoamerican Congress on Pattern Recognition* (pp. 723-731). Springer, Cham. doi: 10.1007/978-3-030-13469-3_84.
[15] Blumenfeld, A., Konen, E., & Greenspan, H. (2018, February). Pneumothorax detection in chest radiographs using convolutional neural networks. In Medical Imaging 2018: Computer-Aided Diagnosis (Vol. 10575, p. 1057504). International Society for Optics and Photonics. doi: 10.1117/12.2292540.
[16] Geva, O., Zimmerman-Moreno, G., Lieberman, S., Konen, E., & Greenspan, H. (2015, March). Pneumothorax detection in chest radiographs using local and global texture signatures. In Medical Imaging 2015: Computer-Aided Diagnosis (Vol. 9414, p. 94141P). International Society for Optics and Photonics. doi:10.1117/12.2083128.
[17] Jakhar, K., Bajaj, R., & Gupta, R. (2019). Pneumothorax Segmentation: Deep Learning Image Segmentation to predict Pneumothorax. *arXiv preprint* arXiv: 1912.07329.
[18] Jun, T. J., Kim, D., & Kim, D. (2018). Automated diagnosis of pneumothorax using an ensemble of convolutional neural networks with multi-sized chest radiography images. *arXiv preprint* arXiv: 1804.06821.
[19] Raghuwanshi, B. S., & Shukla, S. (2019). Class imbalance learning using UnderBagging based kernelized extreme learning machine. *Neurocomputing*, *329*, 172-187. doi: 10.1016/j.neucom.2018.10.056.
[20] Salehinejad, H., Valaee, S., Dowdell, T., Colak, E., & Barfett, J. (2018, April). Generalization of deep neural networks for chest pathology classification in x-rays using generative adversarial networks. In *2018 IEEE International Conference on Acoustics, Speech and Signal Processing (ICASSP)* (pp. 990-994). IEEE. doi: 10.1109/ICASSP.2018.8461430.
[21] Buda, M., Maki, A., & Mazurowski, M. A. (2018). A systematic study of the class imbalance problem in convolutional neural networks. *Neural Networks*, *106*, 249-259. doi: 10.1016/j.neunet.2018.07.011.
[22] He, H., & Garcia, E. A. (2009). Learning from imbalanced data. *IEEE Transactions on knowledge and data engineering*, *21*(9), 1263-1284. doi: 10.1109/TKDE.2008.239.
[23] Raskutti, B., & Kowalczyk, A. (2004). Extreme re-balancing for SVMs: a case study. *ACM Sigkdd Explorations Newsletter*, *6*(1), 60-69. doi: 10.1145/1007730.1007739.
[24] Haixiang, G., Yijing, L., Shang, J., Mingyun, G., Yuanyue, H., & Bing, G. (2017). Learning from class-imbalanced data: Review of methods and applications. *Expert Systems with Applications*, *73*, 220-239. doi: 10.1016/j.eswa.2016.12.035.
[25] Drummond, C., & Holte, R. C. (2003, August). C4. 5, class imbalance, and cost sensitivity: why under-sampling beats over-sampling. In *Workshop on learning from imbalanced datasets II* (Vol. 11, pp. 1-8). Washington DC: Citeseer.
[26] Levi, G., & Hassner, T. (2015). Age and gender classification using convolutional neural networks. In *Proceedings of the IEEE conference on computer vision and pattern recognition workshops* (pp. 34-42).
[27] Chawla, N. V., Bowyer, K. W., Hall, L. O., & Kegelmeyer, W. P. (2002). SMOTE: synthetic minority over-sampling technique. *Journal of artificial intelligence research*, *16*, 321-357. doi: 10.1613/jair.953.

[28] Jo, T., & Japkowicz, N. (2004). Class imbalances versus small disjuncts. *ACM Sigkdd Explorations Newsletter*, *6*(1), 40-49. doi: 10.1145/1007730.1007737.

[29] Guo, H., & Viktor, H. L. (2004). Learning from imbalanced data sets with boosting and data generation: the databoost-im approach. *ACM Sigkdd Explorations Newsletter*, *6*(1), 30-39. doi: 10.1145/1007730.1007736.

[30] Chollet, F. (2016). Building powerful image classification models using very little data [Blog]. Retrieved from https://blog.keras.io/building-powerful-image-classification-models-using-very-little-data.html.

[31] Liu, X. Y., Wu, J., & Zhou, Z. H. (2008). Exploratory undersampling for class-imbalance learning. *IEEE Transactions on Systems, Man, and Cybernetics, Part B (Cybernetics)*, *39*(2), 539-550. doi: 10.1109/TSMCB.2008.2007853.

[32] Sapp, S., van der Laan, M. J., & Canny, J. (2014). Subsemble: an ensemble method for combining subset-specific algorithm fits. *Journal of applied statistics*, *41*(6), 1247-1259. doi: 10.1080/02664763.2013.864263.

[33] Sun, Z., Song, Q., Zhu, X., Sun, H., Xu, B., & Zhou, Y. (2015). A novel ensemble method for classifying imbalanced data. *Pattern Recognition*, *48*(5), 1623-1637. doi: 10.1016/j.patcog.2014.11.014.

[34] Salunkhe, U. R., & Mali, S. N. (2016). Classifier ensemble design for imbalanced data classification: a hybrid approach. *Procedia Computer Science*, *85*, 725-732. doi: 10.1016/j.procs.2016.05.259.

[35] Marouf, M., Siddiqi, R., Bashir, F., & Vohra, B. (2020, January). Automated Hand X-Ray Based Gender Classification and Bone Age Assessment Using Convolutional Neural Network. In *2020 3rd International Conference on Computing, Mathematics and Engineering Technologies (iCoMET)* (pp. 1-5). IEEE. doi: 10.1109/iCoMET48670.2020.9073878.

[36] CS231n Convolutional Neural Networks for Visual Recognition. Retrieved 8 September 2020, from https://cs231n.github.io/transfer-learning/.

[37] Bunrit, S., Kerdprasop, N., & Kerdprasop, K. (2019). Evaluating on the Transfer Learning of CNN Architectures to a Construction Material Image Classification Task. Int. J. Mach. Learn. Comput, 9(2), 201-207. doi: 10.18178/ijmlc.2019.9.2.787.

[38] Simonyan, K., & Zisserman, A. (2014). Very deep convolutional networks for large-scale image recognition. *arXiv preprint* arXiv: 1409.1556.

[39] Mateen, M., Wen, J., Song, S., & Huang, Z. (2019). Fundus image classification using VGG-19 architecture with PCA and SVD. *Symmetry*, *11*(1), 1. doi: 10.3390/sym11010001.

[40] Huang, G., Liu, Z., Van Der Maaten, L., & Weinberger, K. Q. (2017). Densely connected convolutional networks. In *Proceedings of the IEEE conference on computer vision and pattern recognition* (pp. 4700-4708).

[41] Patel, S. (2017). Chapter 2 : SVM (Support Vector Machine) — Theory. Retrieved 7 June 2020, from https://medium.com/machine-learning-101/chapter-2-svm-support-vector-machine-theory-f0812effc72.

[42] Rashid, R., Khawaja, S. G., Akram, M. U., & Khan, A. M. (2018, December). Hybrid RID Network for Efficient Diagnosis of Tuberculosis from Chest X-rays. In *2018 9th Cairo International Biomedical Engineering Conference (CIBEC)* (pp. 167-170). IEEE. doi: 10.1109/CIBEC.2018.8641816.

[43] Scikit-learn: Machine Learning in Python, Pedregosa *et al.*, JMLR 12, pp. 2825-2830, 2011, from https://scikit-learn.org/stable/modules/generated/sklearn.svm.SVC.html.

[44] Marsh (2020). *Chest X-Ray Images with Pneumothorax Masks,* Version 2. Retreived from https://www.kaggle.com/vbookshelf/pneumothorax-chest-xray-images-and-masks.

[45] Random Sample of NIH Chest X-ray Dataset. (2017). Retrieved 6 September 2020, from https://www.kaggle.com/nih-chest-xrays/sample.

[46] Faruqe, M. O., & Hasan, M. A. M. (2009, August). Face recognition using PCA and SVM. In *2009 3rd International Conference on Anti-counterfeiting, Security, and Identification in Communication* (pp. 97-101). IEEE. doi: 10.1109/ICASID.2009.5276938.

[47] Wu, J. D., & Liu, C. T. (2011). Finger-vein pattern identification using SVM and neural network technique. *Expert Systems with Applications*, *38*(11), 14284-14289. doi: 10.1016/j.eswa.2011.05.086.

[48] Priya, R., & Aruna, P. (2012). SVM and neural network based diagnosis of diabetic retinopathy. *International Journal of Computer Applications*, *41*(1). doi: 10.5120/5503-7503

[49] da Silva Santos, M., Ladeira, M., Van Erven, G. C., & da Silva, G. L. (2019, December). Machine Learning Models to Identify the Risk of Modern Slavery in Brazilian Cities. In 2019 18th IEEE International Conference On Machine Learning And Applications (ICMLA) (pp. 740-746). IEEE. doi: 10.1109/ICMLA.2019.00132.

[50] Bharati, S., Podder, P., & Mondal, M. R. H. (2020). Hybrid deep learning for detecting lung diseases from X-ray images. Informatics in Medicine Unlocked, 20, 100391. doi: 10.1016/j.imu.2020.100391

[51] Degiorgis, M., Gnecco, G., Gorni, S., Roth, G., Sanguineti, M., & Taramasso, A. C. (2012). Classifiers for the detection of flood-prone areas using remote sensed elevation data. Journal of hydrology, 470, 302-315. doi: 10.1016/j.jhydrol.2012.09.006.

[52] Mondal, S., Agarwal, K., & Rashid, M. (2019, November). Deep Learning Approach for Automatic Classification of X-Ray Images using Convolutional Neural Network. In 2019 Fifth International Conference on Image Information Processing (ICIIP) (pp. 326-331). IEEE. doi: 10.1109/iciip47207.2019.8985687.